\documentclass[letterpaper, 10 pt, conference]{ieeeconf}  %

\IEEEoverridecommandlockouts                              %

\usepackage[noadjust]{cite}

\usepackage{times}
\usepackage{multicol}

\usepackage{color,soul}
\usepackage{graphicx}
\usepackage{wrapfig}
\usepackage{siunitx}
\usepackage{bbold}
\usepackage{subcaption}
\usepackage{dsfont}
\usepackage{xspace}
\usepackage[font={footnotesize}]{caption} 
\usepackage{amsmath}
\usepackage{bm}
\usepackage{hyperref}
\usepackage{varwidth}

\usepackage{array} %
\usepackage{tabularx}
\usepackage{multirow}
\usepackage{multicol}
\usepackage{booktabs}

\usepackage{algorithm}
\usepackage[noend]{algpseudocode}
\usepackage{amssymb}

\usepackage{amsmath}
\usepackage{bbm}
\usepackage{array}

\newcolumntype{C}{>{\footnotesize}c}

\newcommand{\sysName}{DexMimicGen\xspace}
\newcommand{\sysShort}{DexMG\xspace}

\title{\LARGE \bf
DexMimicGen: Automated Data Generation for\\Bimanual Dexterous Manipulation via Imitation Learning
}

\author{Zhenyu Jiang$^{*1,2}$ \quad Yuqi Xie$^{*1,2}$ \quad Kevin Lin$^{*1,2}$ \quad  \quad Zhenjia Xu$^{1}$ \quad Weikang Wan$^{3}$ \\ \quad Ajay Mandlekar$^{\dag 1}$ \quad Linxi ``Jim" Fan$^{\dag 1}$ \quad Yuke Zhu$^{\dag 1,2}$ 
\thanks{*Equal Contributions, $\dag$Project Leads}%
\thanks{$^{1}$NVIDIA Research, $^{2}$UT Austin, $^{3}$UC San Diego}
}

\begin{document}

\maketitle
\thispagestyle{empty}
\pagestyle{empty}

\begin{abstract}

Imitation learning from human demonstrations is an effective means to teach robots manipulation skills. But data acquisition is a major bottleneck in applying this paradigm more broadly, due to the high costs and human efforts involved. There has been significant interest in imitation learning for bimanual dexterous robots, like humanoids. Unfortunately, data collection is even more challenging here due to the difficulty of simultaneously controlling the two arms and multi-fingered hands. Automated data generation in simulation is a compelling, scalable alternative to fuel this need for training data. To this end, we introduce \sysName, a large-scale automated data generation system that synthesizes trajectories from a handful of human demonstrations for bimanual robots with dexterous hands. We present a collection of simulation environments in the setting of bimanual dexterous manipulation, spanning a range of manipulation behaviors and different requirements for coordination among the two arms. 
We generate 21K demos across these tasks from just 60 source human demos and study the effect of several data generation and policy learning decisions on agent performance.
Finally, we present a real-to-sim-to-real pipeline and deploy it on a real-world humanoid can sorting task. Generated datasets, simulation environments and additional results are at \textcolor{cyan}{\href{https://dexmimicgen.github.io/}{dexmimicgen.github.io}}.

\end{abstract}

\section{Introduction}
\label{sec:intro}

Imitation learning from human demonstrations is an effective means to teach robots manipulation skills~\cite{robomimic2021, brohan2022rt}. 
One popular approach to collecting demonstrations is teleoperation, where human operators control robot arms to collect data for training the autonomous policies~\cite{zhang2017deep, mandlekar2018roboturk}.
Recent efforts have scaled this approach to collect large diverse datasets through teams of human operators, and shown that robots trained on this data can achieve impressive performance and even generalize to different settings~\cite{ebert2021bridge, brohan2022rt, ahn2022can, jang2022bc, lynch2022interactive}. There has also been recent interest in applying this paradigm to humanoid robot embodiments~\cite{darvish2023teleoperation, ding2024bunny, cheng2024open, he2024learning, he2024omnih2o, fu2024humanplus}.

Nonetheless, data acquisition has been a key bottleneck in applying this paradigm more broadly. 
Prior efforts for data collection in the single robot arm setting required multiple human operators, robots, and months of human effort~\cite{ebert2021bridge, brohan2022rt, ahn2022can, jang2022bc, lynch2022interactive}. Unfortunately, scaling data collection for humanoids can be even more difficult, owing to the challenges of controlling the two arms and multi-fingered dexterous hands simultaneously. Enabling real-time teleoperation for humanoids has required the development of special-purpose teleoperation interfaces~\cite{darvish2023teleoperation, ding2024bunny, cheng2024open, he2024learning, he2024omnih2o, fu2024humanplus}, but these pipelines can be costly and difficult to scale. Furthermore, the increase in operator burden due to multi-arm and multi-finger hand control makes collecting demonstrations in this setting more challenging compared to the single-arm setting, further limiting the rate of data collection. The data acquisition burden is further compounded by the higher data requirements in the humanoid setting due to the increased degrees of freedom and task complexity.

Leveraging automated data generation in simulation is a compelling alternative that has proved effective for the single-arm robot manipulation setting~\cite{dalal2023imitating, wang2023robogen, mandlekar2023mimicgen}. 
Inspired by prior successes,  \textbf{we introduce \sysName (\sysShort), a large-scale automated data generation system for bimanual robots with dexterous hands, such as humanoids}. The core idea is to leverage a small set of human demonstrations and use demonstration transformation and replay in physical simulation to automatically generate large amounts of training data suitable for imitation learning in the bimanual dexterous manipulation setting. 
This system builds on top of MimicGen~\cite{mandlekar2023mimicgen}, which proposed a similar pipeline for the single-arm with parallel-jaw gripper setting. However, there remain several technical challenges that \sysName has to overcome to operationalize the same principles.

\begin{figure}[t!]
    \centering
    \includegraphics[width=\linewidth]{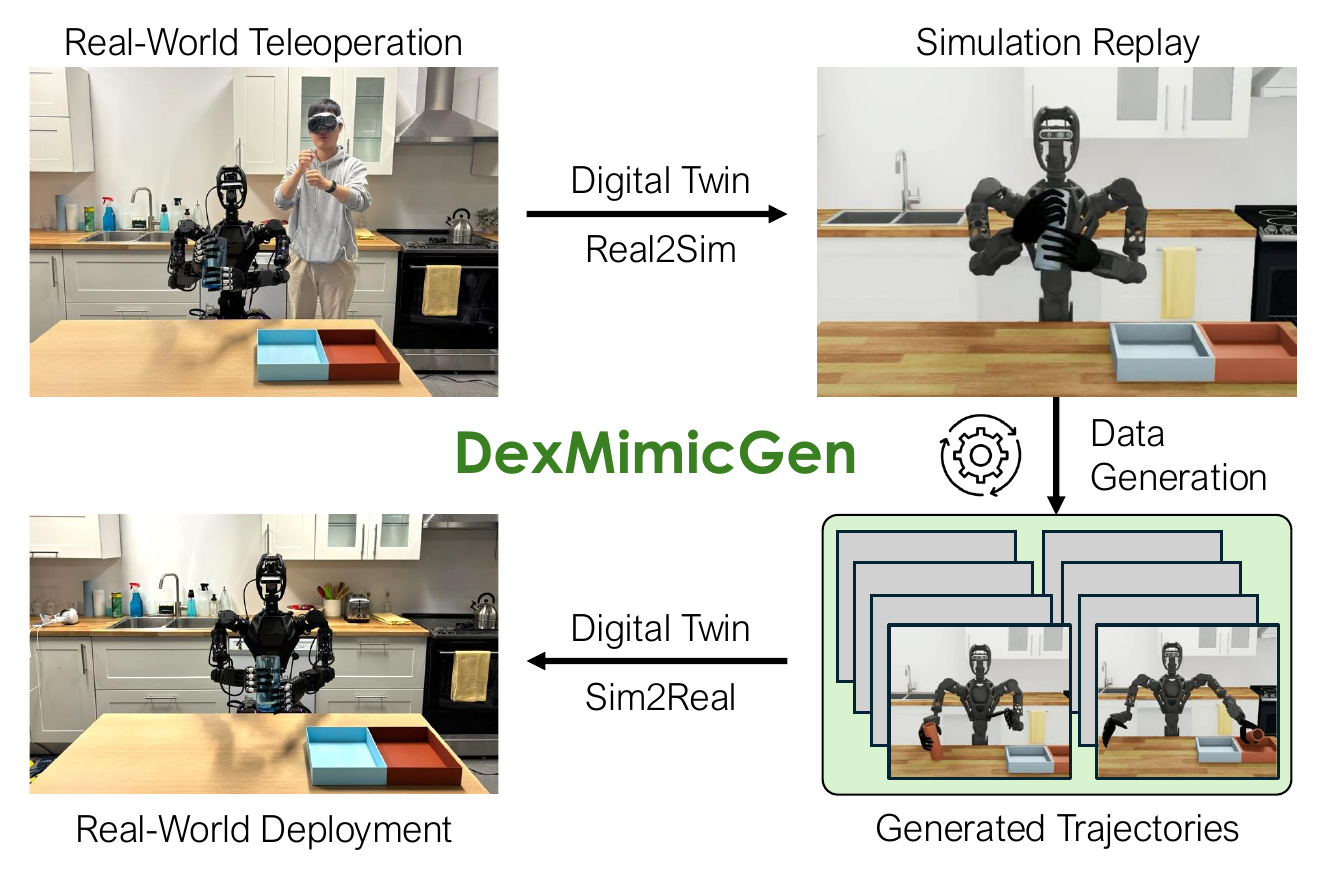}
    \caption{\textbf{\sysName Overview.} \sysName offers an efficient pipeline to train capable bimanual dexterous robots. (left) First, a human operator collects around five task demonstrations using a teleoperation device. (middle) Next, \sysName automatically generates a large set of demonstration trajectories in simulation. (right)  Finally, a policy is trained with imitation learning and deployed in the real world.
    }
    \label{fig:pull}
    \vspace{-16pt}
\end{figure}

MimicGen relies on decomposing each task into a sequence of subtasks, to generate trajectories for each subtask separately and then stitch them together.
Bimanual dexterous manipulation involves three types of subtasks where the two arms need to achieve sub-goals independently, with coordination, and following a specific order. MimicGen, which relies on a single subtask segmentation, struggles to handle the independent and interdependent actions required in bimanual tasks.
To address these challenges, \sysName incorporates a flexible per-arm subtask segmentation strategy, allowing each arm to execute its subtasks independently while still accommodating the necessary coordination phases. 
\sysShort employs a synchronization strategy to ensure precise alignment of actions during coordination subtasks and an ordering constraint mechanism to enforce the correct order of actions during sequential subtasks.

\vspace{1mm}
\noindent \textbf{We make the following contributions:}
\newline\noindent $\bullet$ We introduce \sysName (\sysShort), a data generation system that automatically synthesizes trajectories from a small number of human demonstrations for bimanual and dexterous robot manipulation. We introduce several key design features, including an asynchronous per-arm execution strategy, synchronization, and sequential constraints that enable handling multi-arm coordination.
\newline\noindent $\bullet$ We introduce a suite of nine simulation environments across three different embodiment types 
requiring different coordination behaviors between the two arms.
We apply \sysName to generate 21K demos across these tasks from merely 60 source human demos and study the effect of several data generation and policy learning decisions on agent performance. We have released our simulations and datasets to facilitate future study into the bimanual and dexterous manipulation setting.
\newline\noindent $\bullet$ We create a simulated digital twin of a real-world can-sorting task, replay real-world human demonstrations in the simulation, synthesize trajectories with \sysName, and then transfer the generated trajectories back into the real world, producing a visuomotor policy of $90\%$ success rate, as opposed to $0\%$ from just using the human demos (Fig.~\ref{fig:pull}).

\section{Related Work}
\label{sec:related_work}

\textbf{Data Collection through Teleoperation.} 
Teleoperation is a prevalent approach to gathering task demonstrations in robotics~\cite{zhang2017deep, mandlekar2018roboturk, mandlekar2019scaling, mandlekar2020human, wong2022error, wu2023gello, iyer2024open, dass2024telemoma}. Human operators use an interface to control a robot in real time remotely, and sensor data and robot control commands are logged to a dataset.
Some systems allow data collection for multiple robot arms~\cite{tung2020learning, zhao2023learning, aldaco2024aloha,lin2024learning} and humanoids~\cite{schaal1999imitation, darvish2023teleoperation, ding2024bunny, cheng2024open, he2024learning, he2024omnih2o, fu2024humanplus}, and some also enable robot-free data collection using specialized hardware~\cite{fang2023low, chi2024universal, etukuru2024robot}.
However, all these methods require significant human time and resources to collect large datasets.
Some other efforts use pre-programmed experts to automate data generation in simulation~\cite{james2020rlbench, zeng2020transporter, jiang2022vima, gu2023maniskill2, dalal2023imitating, ha2023scaling, wang2023robogen}, but applying these methods to challenging scenarios involving multi-arm coordination can be difficult.
By contrast, \sysName builds upon MimicGen~\cite{mandlekar2023mimicgen, hoque2023interventional, robocasa2024} to automate data generation using a handful of human demonstrations, greatly reducing the human effort involved in collecting large datasets. 

\textbf{Imitation Learning and Data Augmentation.}
Behavioral Cloning~\cite{pomerleau1989alvinn} is an established framework for learning robot control policies from demonstrations and has been used extensively in prior work~\cite{finn2017one, Billard2008RobotPB, Calinon2010LearningAR, mandlekar2020learning, zeng2020transporter, wang2021generalization, lynch2019learning,pertsch2021skild,ajay2020opal,hakhamaneshi2021fist,zhu2022bottom,nasiriany2022sailor}, including for bimanual manipulators~\cite{drolet2024, tung2020learning, zhao2023learning, aldaco2024aloha} and humanoid robots~\cite{schaal1999imitation, Ijspeert2002MovementIW, seo2023deep, ding2024bunny, cheng2024open}.
In this work, we apply existing imitation learning methods~\cite{robomimic2021, chi2023diffusion} to datasets generated by \sysName. We show \sysName plays a significant role in facilitating algorithm development for bimanual manipulation by making simulation-based manipulation datasets more widely accessible and providing easy-to-reproduce results.
Recent works have leveraged offline data augmentation to increase the dataset sizes~\cite{mitrano2022data, laskin2020reinforcement, kostrikov2020image, young2020visual, zhan2020framework, robomimic2021, sinha2022s4rl, pitis2020counterfactual, pitis2022mocoda, mandi2022cacti, yu2023scaling, chen2023genaug, bharadhwaj2023roboagent, zhang2024diffusion, tian2024vista, chen2024roviaug}. By contrast, \sysName generates datasets using online simulation, ensuring the generated trajectories are physically valid.

\section{Prerequisites}
\label{sec:problem}

\textbf{Imitation Learning.} We formalize each manipulation task as a Partially Observable Markov Decision Process (POMDP). We are given $N$ demonstrations $\mathcal{D} = \{(s_0^i, o_0^i, a_0^i, s_1^i, o_1^i, a_1^i, ..., s_{H_i}^i)\}_{i=1}^N$ consisting of states $s \in {\cal S}$, observations $o \in {\cal O}$, and actions $a \in {\cal A}$. Each episode starts in a state $s_0^i \sim D$ sampled from the initial state distribution $D \subseteq {\cal S}$. The goal is to learn a policy $\pi: {\cal O} \to {\cal A}$ that maps observations to a distribution over the action space. We focus on Behavioral Cloning (BC)~\cite{pomerleau1989alvinn} methods that find a policy via the maximum likelihood objective $\arg\max_{\theta} \mathbb{E}_{(s, o, a) \sim \mathcal{D}} [\log \pi_{\theta}(a \mid o)]$. We train our policies with datasets generated via \sysName.

\textbf{Assumptions.} Like MimicGen~\cite{mandlekar2023mimicgen}, we make these assumptions. (\textbf{A1}): the action space $\mathcal{A}$ consists of the following components for each robot arm: a pose command for an end effector controller and an actuation command for the hand ($1$-D open/close for parallel-jaw gripper, $6$-D joint commands for dexterous hand). (\textbf{A2}): Each task can be divided into object-centric subtasks (see Sec.~\ref{subsec:parallel}). (\textbf{A3}): During data collection, an object's pose can be observed or estimated prior to a robot arm making contact with that object.

\textbf{MimicGen.} 
MimicGen~\cite{mandlekar2023mimicgen} uses a small number of source human demonstrations $\mathcal{D}_{\text{src}}$ to generate a large dataset $\mathcal{D}$. 
It assumes that every task consists of a sequence of object-centric subtasks ($S_1(o_1)$, $S_2(o_2)$, \ldots, $S_M(o_M)$) where the manipulation in each subtask $S_i(o_i)$ is relative to a single object's coordinate frame ($o_i \in \mathcal{O}$, where $\mathcal{O}$ is the set of objects in the task).
It divides each source demo $\tau \in D_{src}$ into contiguous object-centric manipulation segments $\{\tau_i\}_{i=1}^M$, each of which corresponds to a subtask $S_i(o_i)$.
Each segment is a sequence of end effector control poses $\tau_i = (T^{C_0}_W, T^{C_1}_W, ..., T^{C_K}_W)$ where $W$ is the world reference frame.
This segmentation can be done with human annotation or using heuristics.
To generate a new demonstration in a novel scene, it observes the pose of the object for the current subtask $T^{o'_i}_{W}$, and transforms the poses in a source human segment (with a constant SE(3) transform $T^{o'_i}_{W}(T^{o_i}_{W})^{-1}$) such that relative poses between the end effector and object frame are preserved between the source segment and the new scene. 
It then adds poses to the start of the segment to interpolate between the robot's current state and the start of the transformed segment. 
Then, it executes the entire sequence of poses open-loop using the robot end effector controller and repeats this process for the next subtask. It checks for task success after executing all subtasks and only keeps the demonstration if it was successful.

\begin{figure}[t!]
    \centering
    \vspace{3pt}
    \includegraphics[width=\linewidth]{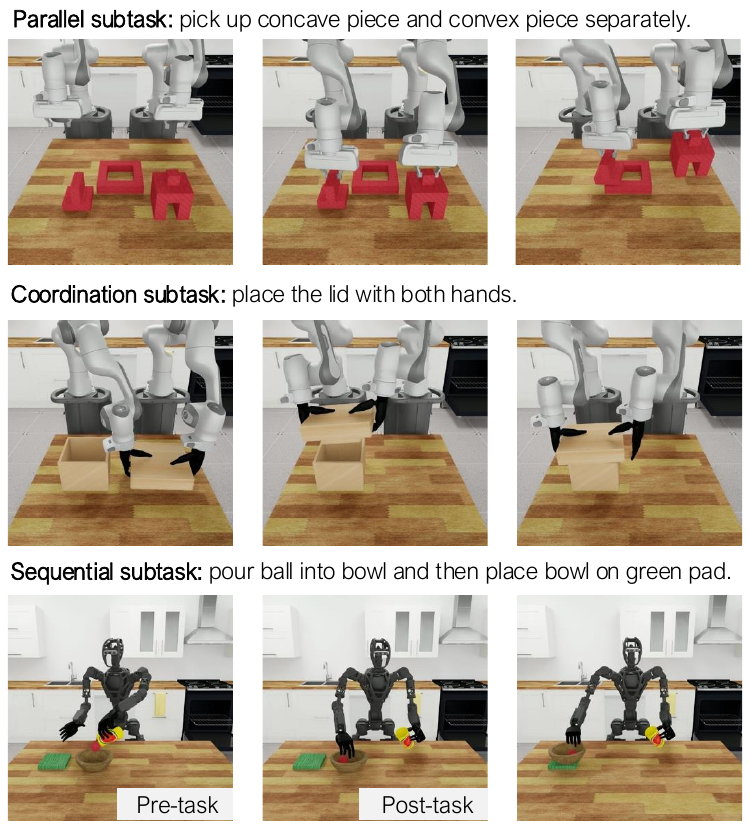}
    \caption{\textbf{Subtask Types.} We categorize the subtasks into parallel, coordination, and sequential subtasks, where the two arms achieve subgoals independently, with coordination, and following a specific order.}
    \label{fig:subtask}
    \vspace{-18pt}
\end{figure}

\section{\sysName Method}
\label{sec:method}

\sysName generates data for bimanual and dexterous manipulation --- doing so involves handling three key challenges compared to MimicGen. First, each arm must operate independently of the other arm to achieve different goals. Next, the arms must coordinate to accomplish a shared goal. Finally, one arm's subtask must be completed before the next one can be attempted.
\sysName handles these challenges by introducing a taxonomy of subtask types (Fig.~\ref{fig:subtask}) --- parallel (Sec.~\ref{subsec:parallel}), coordination (Sec.~\ref{subsec:coord}), and sequential (Sec.~\ref{subsec:concurrency}), and making changes to the data generation process to accommodate them. Sec.~\ref{subsec:datagen_bimanual} provides an overview of the entire data generation process.
Note that, similar to MimicGen, we exploit the SE(3) equivariance of robot actions with respect to object poses. Specifically, when an object's pose has an SE(3) transformation applied to it, we can similarly apply the same SE(3) transformation to robot actions to replicate the same effect of the original robot actions on the new object pose.

\begin{figure*}[t!]
    \centering
    
    \vspace{4pt}
    \includegraphics[width=\linewidth]{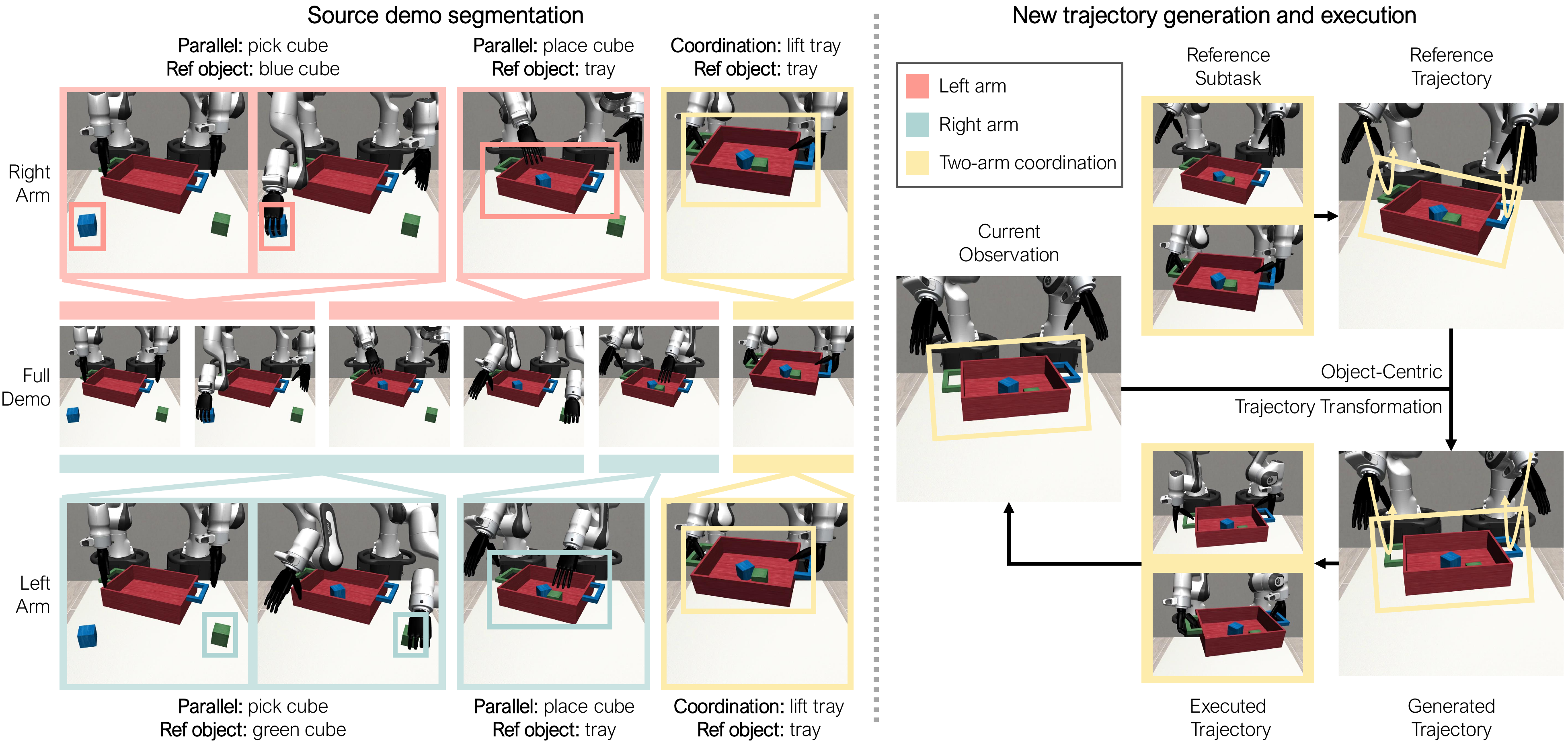}
    \caption{\textbf{\sysName Workflow.} Left: segment source demonstrations for each arm through manually defined heuristics or human and records the poses of the reference objects. Right: In a new simulation environment, we generate trajectories by transforming source trajectories with reference object poses and executing them.}
    \label{fig:pipeline}
    \vspace{-18pt}
\end{figure*}

\subsection{Parallel Subtasks}
\label{subsec:parallel}

In the bimanual setting, each arm must be able to operate independently of the other arm. For example, at the start of the Piece Assembly task (Fig.~\ref{fig:subtask} top), each arm needs to grasp a separate object and might finish grasping the object at different points in time. This makes the single fixed sequence of subtasks from MimicGen unsuitable.
To enable a flexible order of completion for \textbf{parallel subtasks} involving two arms, we consider each task to consist of a sequence of subtasks for each arm: $S^{a_1}_1(o_1)$, ..., $S^{a_1}_{M_1}(o_{M_1})$ and $S^{a_2}_1(o_1)$, ..., $S^{a_2}_{M_2}(o_{M_2})$.
Each source demonstration is split into object-centric manipulation segments as in MimicGen, but now each arm has its own set of segments ($\{\tau^n_i\}_{i=1}^{M_n}$, $n \in \{1, 2\}$).

However, since arm subtasks are defined independently, their execution can start and end at different times that are not aligned. To accommodate this, \sysName employs an asynchronous execution strategy, where an action queue is maintained for each arm. Actions are dequeued for each arm one by one in parallel.
Whenever an arm's queue is empty, it is populated with the transformed subtask segment for the next subtask (using the same transformation from MimicGen). This approach allows for the execution of actions for both arms without requiring alignment between subtasks.

\subsection{Coordination Subtasks}
\label{subsec:coord}

Some tasks require precise coordination, such as placing the lid in the Box Cleanup task (Fig.~\ref{fig:subtask} middle). In these \textbf{coordination subtasks}, the relative poses between the two end-effectors during execution must be aligned with the corresponding relative poses in the source demonstration. To achieve this, we ensure that 1) both arms execute their trajectories in a synchronized manner and 2) the trajectories for both arms are generated with the same transformation.
To achieve this temporal alignment, 
we enforce that coordination subtasks end at the same timestep during source demo segmentation.
During execution, we implement a synchronization strategy in which each arm waits for the other until both have the same number of remaining steps in the coordination subtask, aligning the end of subtask execution with the subtask segmentation.

We provide different source demonstration transformation schemes to acquire the common transformation matrix for both arms in coordination subtasks. These include the Transform and Replay schemes. The Transform scheme utilizes the transformation matrix $T^{o'_i}_{W}(T^{o_i}_{W})^{-1}$ computed from the object pose at the moment the first arm begins the coordination subtask $T^{o'_i}_{W}$ and the object pose in the corresponding source segment $T^{o_i}_{W}$. In contrast, the replay scheme directly uses the source trajectories without applying any transformation. The replay scheme can be beneficial for specific coordination subtasks like the handover phase of the Can Sorting and Transport tasks, because it ensures the trajectory remains within kinematic limits and is fully executable.

\subsection{Sequential Subtasks}
\label{subsec:concurrency}

Some tasks require subtasks to be completed in a specific order. 
For example, in the Pouring task (Fig.~\ref{fig:subtask} bottom), the robot must pour the ball into the bowl with one hand before moving the bowl to the pad with the other hand. To handle these \textbf{sequential subtasks}, we implement an ordering constraint mechanism. We specify a pre-subtask (pouring the ball) and a post-subtask (picking the bowl) based on the task requirement. This mechanism ensures that the arm executing the post-subtask waits until the pre-subtask of the other arm is completed before continuing with the post-subtask.

\subsection{Data Generation for Bimanual Manipulation}
\label{subsec:datagen_bimanual}

We outline the overall \sysName data generation workflow using the Tray Lift task as an example.
First, source demos are segmented into per-arm subtasks using manually defined heuristics or human annotation (Fig.~\ref{fig:pipeline} left).
The final subtask for each arm requires coordination (they must lift the tray together), so it is annotated as a coordination subtask for synchronization during data generation (Sec.~\ref{subsec:coord}).

At the start of data generation, the scene is randomized and a source demonstration is selected (as in MimicGen). 
We then iteratively generate and execute trajectories for each subtask of each arm in parallel (see Fig.~\ref{fig:pipeline} right). 
In this example, given the pose of the reference object (the tray), we compute the relative transformation between the current tray pose and the tray pose in the source segment. We use this transformation to transform the source trajectories of both arms because these are coordination subtasks. Then we use the synchronization execution strategy described in Sec.~\ref{subsec:coord} to execute the generated trajectory.
Note that we generate finger motion by replaying the finger joint actions in the source demo because the finger movement is always relative to the end-effector movement. 
Each generated demonstration is only kept if the task is successful, and this process repeats until a sufficient amount of data is generated.

\section{System Design}
\label{sec:setup}

In order to instantiate \sysName, we build a large collection of simulation environments and a teleoperation system allowing for source human demonstration collection in both simulation and the real world.

\textbf{Simulation Environments.}
We introduce a diverse range of setups and tasks to demonstrate the capability of \sysName to generate data across different embodiments and manipulation behaviors. 
The tasks are developed in RoboSuite~\cite{robosuite2020} and use MuJoCo~\cite{todorov2012mujoco} for physics simulation.
We focus on three embodiments: (1) bimanual Panda arms equipped with parallel-jaw grippers, (2) bimanual Panda arms with dexterous hands, and (3) a GR-1 humanoid equipped with dexterous hands. We apply different controllers for different embodiments. For the Panda arms, we leverage the Operational Space Control (OSC) \cite{khatib1987unified} framework, which converts the delta end-effector pose into joint torque commands. For the humanoid, we implemented an Inverse Kinematics (IK) controller based on mink~\cite{mink2024, pink2024}. We found this to be an effective approach to deal with the complexity of the humanoid kinematic tree, where both arms are linked to a single torso. The IK controller translates global target end-effector poses into robot joint positions. For finger control, we directly use joint position control.

\begin{figure}[t!]
    \centering
    \vspace{6pt}
    \includegraphics[width=\linewidth]{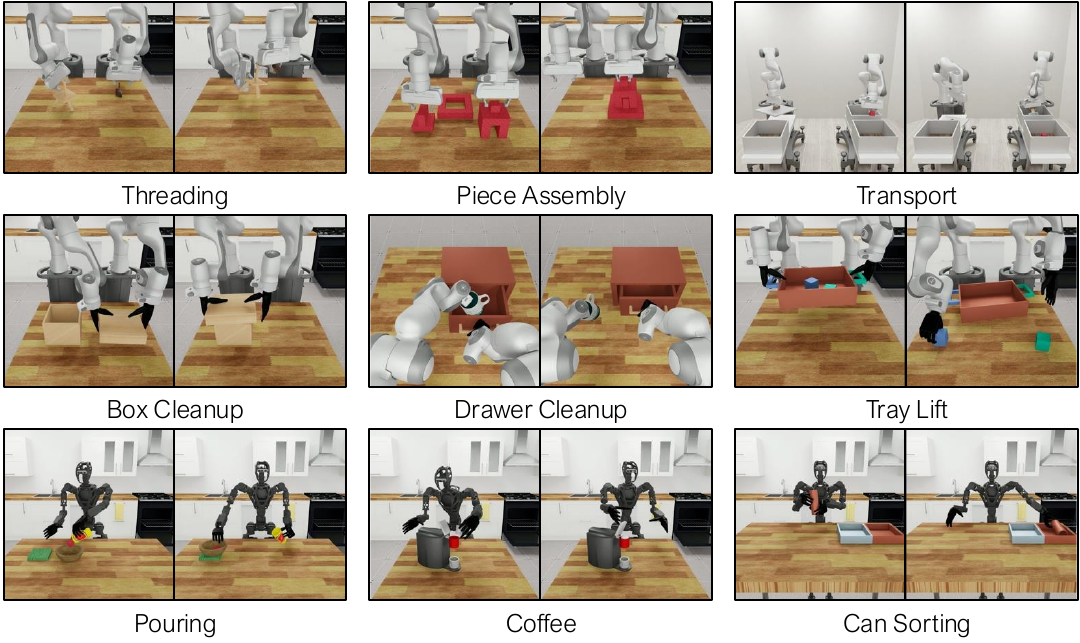}
    \caption{\textbf{Simulation Tasks.} We deploy \sysName on nine simulation tasks across three embodiments --- two arms with parallel-jaw grippers (top), two arms with dexterous hands (middle), and a humanoid (bottom)}
    \label{fig:tasks_sim}
    \vspace{-10pt}
\end{figure}

For each embodiment, we introduce three tasks, resulting in a total of nine tasks, as depicted in Fig.~\ref{fig:tasks_sim}. These tasks involve high-precision manipulation (Threading, Piece Assembly, Box Packing, Coffee), manipulation of articulated objects (Drawer), and are long-horizon (Transport).
The tasks also require overcoming key challenges in multi-arm interaction.
Several of these tasks contain coordination subtasks, where both arms need to cooperate to finish the subtask (Threading, Transport, Box Packing, Tray Lift, Can Sorting). 
Other tasks necessitate sequential subtask execution (Piece Assembly, Drawer Cleanup, Pouring, Coffee). 
We also introduce task variants that broaden the default reset distribution $D_0$ for certain tasks, as in MimicGen.
For instance, in the Pouring task, $D_1$ represents a variant where objects have a larger initial reset distribution, while in $D_2$, the reset positions of the bowl and the green pad are swapped.
These simulation environments along with the datasets generated by \sysName provide a valuable platform to analyze various factors that influence the performance of imitation learning in the bimanual and dexterous manipulation setting.

\textbf{Teleoperation System.}
To collect source demonstrations for the tasks, we employ different teleoperation methods tailored to each embodiment. For bimanual Panda arms equipped with parallel-jaw grippers, we use an iPhone-based teleoperation interface, as introduced in RoboTurk~\cite{mandlekar2018roboturk, tung2020learning}, to capture human wrist and gripper actions. For robots equipped with dexterous hands, we implemented an Apple Vision Pro-based teleoperation system. Specifically, we employ the VisionProTeleop software~\cite{park2024avp} to collect wrist and finger poses via Apple Vision Pro. We first align the human and the robot to convert the raw human end effector poses to robot poses. We design a human-to-robot calibration process asking the human teleoperator to start with a fixed pose, and we automatically compute the relative transformation matrices that map the human poses to robot targets. This calibration process adapts to both bimanual Panda arms with dexterous hands and the GR-1 humanoid. We use the retargeting method provided by OmniH2O~\cite{he2024omnih2o} to retarget human finger pose to robot finger joint positions. This teleoperation system converts human actions to robot action targets, allowing us to collect demonstrations intuitively.

\begin{table}[t!]
\centering

\centering

\vspace{5pt}
\resizebox{1.0\linewidth}{!}{
\begin{tabular}{lcccc}
\toprule
\textbf{Task} & 
\textbf{Source Demo DP} & 
\textbf{BC-RNN-GMM} & 
\textbf{BC-RNN} & 
\textbf{DP} \\ 
\midrule
Piece Assembly  & $3.3 \pm 0.9$ & $74.0 \pm 2.8$ & $66.0 \pm 2.0$ & $\mathbf{80.7 \pm 0.9}$ \\
Threading  & $1.3 \pm 0.9$ & $54.0 \pm 4.3$ & $68.0 \pm 1.6$ & $\mathbf{69.3 \pm 1.9}$ \\
Transport  & $52.7 \pm 7.7$ & $64.0 \pm 3.3$ & $57.3 \pm 4.1$ & $\mathbf{83.3 \pm 0.9}$ \\
Box Cleanup  & $62.0 \pm 1.6$ & $64.0 \pm 10.0$ & $\mathbf{94.7 \pm 0.9}$ & $92.0 \pm 4.3$ \\
Drawer Cleanup  & $0.7 \pm 0.9$ & $30.7 \pm 5.0$ & $\mathbf{80.0 \pm 0.0}$ & $76.0 \pm 0.0$ \\
Tray Lift  & $3.3 \pm 0.9$ & $66.0 \pm 8.2$ & $78.0 \pm 2.0$ & $\mathbf{88.7 \pm 0.9}$ \\
Pouring  & $0.7 \pm 0.9$ & $74.0 \pm 8.6$ & $62.0 \pm 7.5$ & $\mathbf{79.3 \pm 0.9}$ \\
Coffee  & $14.7 \pm 0.9$ & $12.0 \pm 1.6$ & $\mathbf{84.7 \pm 4.9}$ & $77.3 \pm 0.9$ \\
Can Sorting  & $0.7 \pm 0.9$ & $75.3 \pm 1.9$ & $96.0 \pm 4.3$ & $\mathbf{97.3 \pm 0.9}$ \\
\bottomrule
\vspace{-11pt}
\end{tabular}
}

\caption{Success rates (3 seeds) of image-based policies trained with BC on the source demos and \sysName datasets of 1000 trajectories.}
\label{tab:core-results-image}
\vspace{-8pt}
\end{table}

\section{Experiments}
\label{sec:experiments}

In this section, we provide empirical evidence showcasing the efficacy of \sysName. We discuss details on experiment setup (Sec.~\ref{subsec:implementation}), highlight \sysName features and applications (Sec.~\ref{subsec:features}), then analyze how data generation and policy learning choices impact policy performance (Sec.~\ref{subsec:analysis}), and finally present a real-world application of the \sysName system (Sec.~\ref{subsec:real_world}).

\subsection{Experimental Setup}
\label{subsec:implementation}
We collect ten source human demonstrations for each task with parallel-jaw grippers, but only five demonstrations for those involving dexterous hands due to the additional operator burden and time cost of collecting demonstrations for dexterous hands. \sysName is subsequently used to generate 1000 demonstrations per task. Each dataset was used to train visuomotor policies through Behavioral Cloning with an RNN~\cite{robomimic2021}, an RNN-GMM~\cite{robomimic2021}, and a Diffusion Policy~\cite{chi2023diffusion}. For evaluation, we follow the procedure in prior work~\cite{robomimic2021, mandlekar2023mimicgen}: we run each experiment across 3 different seeds, and take the maximum policy success rate for each seed.

\subsection{\sysName Features}
\label{subsec:features}

\begin{figure}[t!]
\centering
\vspace{4pt}
\includegraphics[width=\linewidth]{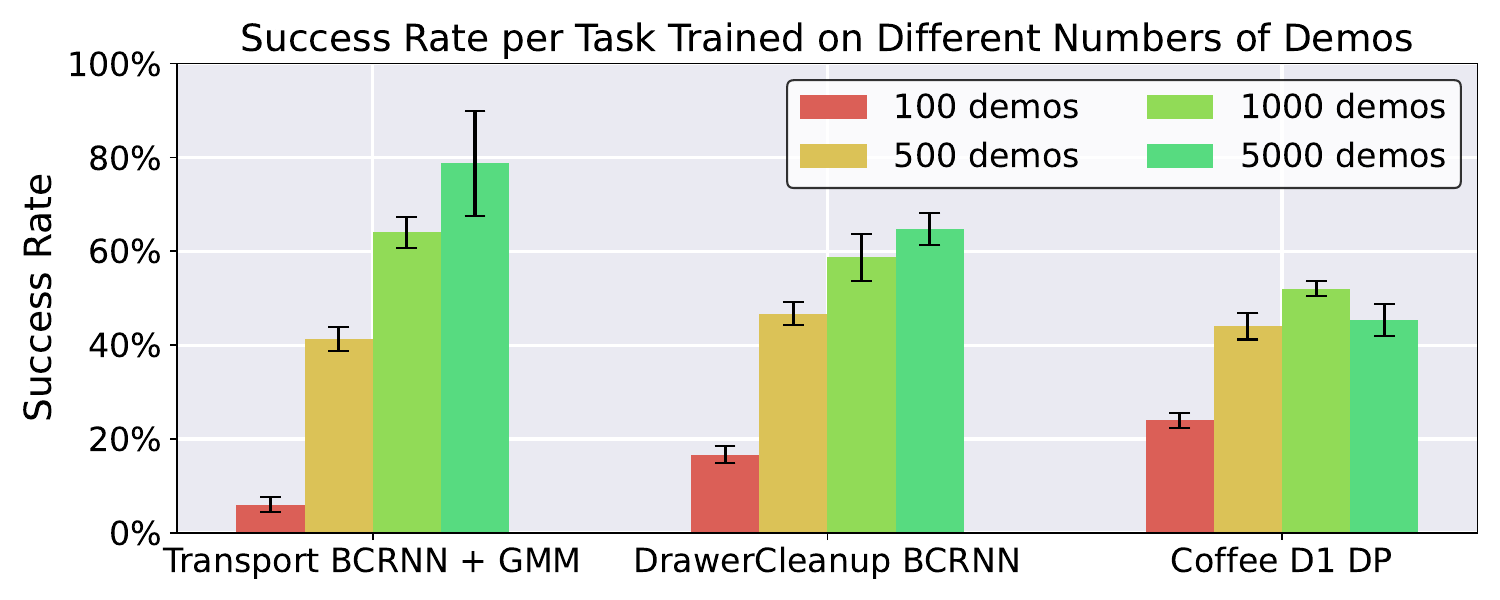}

\caption{\textbf{Dataset Size Comparison.} Success rates of policies trained on datasets with different sizes.} %
\label{fig:scaling-law}
\vspace{-5pt}
\end{figure}

\begin{table}[t]
\vspace{6pt}
\centering
\resizebox{1.0\linewidth}{!}{
\begin{tabular}{llccc}
\toprule
\textbf{Task} & \textbf{Policy} & \textbf{D0} & \textbf{D1} & \textbf{D2} \\ 
\midrule
Piece Assembly & BC-RNN-GMM & $74.0 \pm 2.8$ & $67.3 \pm 0.9$ & $44.0 \pm 3.3$ \\
Box Cleanup & BCRNN & $96.7 \pm 2.5$ & $88.0 \pm 5.9$ & $78.7 \pm 2.5$ \\
Pouring & DP & $79.3 \pm 0.9$ & $82.0 \pm 2.0$ & $71.3 \pm 2.5$ \\
\bottomrule
\end{tabular}
}
\caption{Success rates of policy trained on data generated with broader initial distributions, evaluated with same broader initial distributions.}
\label{tab:large-reset-distr}
\vspace{-18pt}
\end{table}

\begin{table}[t]
\vspace{7pt}
\centering
\resizebox{1.0\linewidth}{!}{
\fontsize{5}{5}\selectfont %
\begin{tabular}{llcc}
\toprule
\textbf{Task} & \textbf{Policy} & \textbf{DemoNoise} & \textbf{DexMimicGen} \\ 
\midrule
Piece Assembly & BCRNN + GMM & 12.7 $\pm$ 3.4 & \textbf{74.0} $\pm$ \textbf{2.8} \\
Tray Lift & BCRNN & 16.7 $\pm$ 2.5 & \textbf{75.3} $\pm$ \textbf{7.5} \\
Pouring & DP & 26.7 $\pm$ 2.5 & \textbf{79.3} $\pm$ \textbf{0.9} \\
\bottomrule
\end{tabular}
}
\caption{Success rates of policies trained on data generated with \sysName and Demo-noise baseline.}
\label{tab:demo-noise-baseline}
\vspace{-18pt}
\end{table}

\textbf{\sysName significantly boosts the policies' success rates over using the source demonstrations only.} Robots trained on \sysName's datasets outperform those trained only on the small source datasets across all tasks (see Table~\ref{tab:core-results-image}).
Notable improvements include policy performance on Drawer Cleanup (0.7\% to 76.0\% success), Threading (1.3\% to 69.3\%), and Piece Assembly (3.3\% to 80.7\%).

\textbf{\sysName produces capable policies across diverse initial state distributions}.
\sysName generates datasets with broader initial state distributions ($D_1$, $D_2$) from source demos in $D_0$. As shown in Table~\ref{tab:large-reset-distr}, policies trained on these datasets are performant in the evaluation with the same broader initial state distributions, showing that \sysName generates valuable datasets on new initial state distributions.

\textbf{\sysName generates data across different benchmarks}.
We apply \sysName to BiGym~\cite{chernyadev2024bigym}, a new simulation benchmark for humanoid robots involving bimanual mobile manipulation tasks. We generate 1000 demonstrations for each of the three tasks, FlipCup, DishwasherLoadPlates, and CupBoardsCloseAll, and achieve data generation success rates of 29.1\%, 43.6\%, and 76.4\%. The visualizations of generated demonstrations can be found on the project website.

\subsection{\sysName Analysis}
\label{subsec:analysis}

\textbf{How does \sysName data generation compare to alternatives?} We compare \sysName with a Demo-Noise data generation baseline, which takes the same source demonstrations as \sysName, but generates data by replaying the source demos with action noise during execution. In Table~\ref{tab:demo-noise-baseline}, we train policies on datasets of 1000 demos generated by both \sysName and the Demo-Noise baseline. We can see that the policies trained using \sysName outperform those trained on the Demo-Noise baseline by more than 58\% across all tasks. Furthermore, unlike \sysName, the Demo-Noise baseline cannot generate results on $D_1$ and $D_2$, as it can only replay the same initial configurations in the source demos.

\textbf{Do larger datasets boost policy performance?} We train policies on 100, 500, 1000, and 5000 demos generated by \sysName across several tasks (Fig. \ref{fig:scaling-law}). 
We observe significant boosts in performance from 100 to 500 and 1000, showing that increasing dataset size boosts performance in this data regime; however, the success rate does not always increase from 1000 to 5000, suggesting that there can be diminishing returns depending on the task.

\textbf{How do different \sysName data generation strategies impact results?} 
First, we compare the Replay and Transform schemes in the coordination subtask (Sec.~\ref{subsec:coord}). Specifically, we evaluate two tasks involving the handover subtask with two distinct policies: Transport using BCRNN+GMM, and Can Sorting using a diffusion policy.
Replay demonstrates better policy performance (63.3\% vs. 46.0\%) in the Transport task and achieves comparable outcomes (97.3\% vs. 98.6\%) in the Can Sorting task. Thus, Replay is our default choice for tasks that involve handover.

Next, we assess the effectiveness of ordering constraints in sequential subtasks (Sec.~\ref{subsec:concurrency}). When using the same source demonstration for both arms, subtask ordering requirements are typically satisfied automatically. In contrast, employing different source demonstrations for each arm requires an ordering constraint but also increases data diversity. We also evaluate two tasks involving the sequential subtasks with two distinct policies: Drawer Cleanup with BCRNN, and Pouring with diffusion policy.
We found training on data generated with ordering constraints consistently outperforms training without them (50.7\% vs. 48.0\% in Drawer Cleanup and 88.7\% vs. 76.7\% in Pouring). Directly using the same source demo yields the policy success rates of 56.7\% in the Drawer Cleanup and 79.3\% in Pouring.

\textbf{How do different policy architecture choices affect success rates?} 
In Table~\ref{tab:core-results-image}, we also compare the performance of different policy architectures (Diffusion Policy~\cite{chi2023diffusion}, BC-RNN-GMM~\cite{robomimic2021}, BC-RNN~\cite{robomimic2021} with no GMM action head) on the datasets generated by \sysName. 
We found that Diffusion Policy \cite{chi2023diffusion} generally outperforms the other architectures. Interestingly, we also found that BC-RNN-GMM generally underperformed BC-RNN and Diffusion Policy, especially on tasks that involve dexterous hands, in contrast to the RoboMimic study~\cite{robomimic2021} which found the use of a GMM head to be beneficial.
We believe \sysName datasets will make it easier for future work to study further how imitation learning choices might differ in the bimanual dexterous manipulation setting.

\begin{figure}[t!]
\vspace{5pt}
    \centering
    \includegraphics[width=\linewidth]{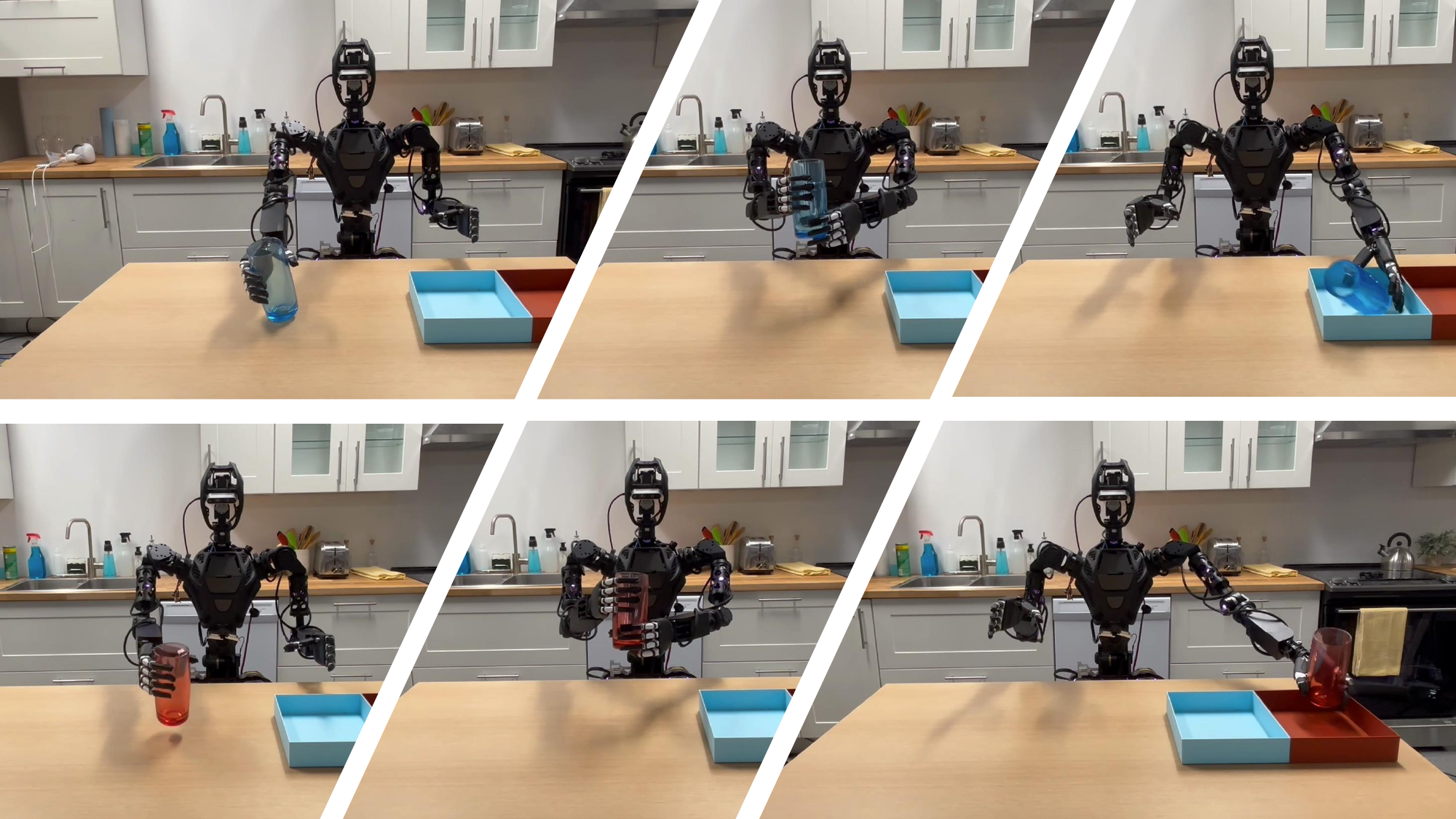}
    \caption{\textbf{Real-World \sysName Deployment.} Rollouts of real-world visuomotor policy trained with \sysName data and digital twin.}
    \label{fig:real}
    \vspace{-18pt}
\end{figure}

\subsection{Real-World Evaluation}
\label{subsec:real_world}

We showcase how \sysName enables real-world deployment using the pipeline illustrated in Fig.~\ref{fig:pull}. We generate real-world demonstrations by running \sysName with a digital twin~\cite{jiang2022ditto} in simulation.

\textbf{Hardware Setup.} We use a Fourier GR1 robot equipped with two 6-DoF Inspire dexterous hands. For vision, we use two Intel RealSense D435i cameras: one head-mounted camera provides a first-person view and one camera in front of the robot as a third-person view. %

\textbf{Digital Twin Setup.} 
We perform our experiment on the Can Sorting task (Fig.~\ref{fig:real}), with digital twin assets in simulation that align with the real-world setup. 
To ensure accurate alignment between the real-world and simulated environments, we perform pose estimation on the objects prior to data collection. Using the head-mounted camera, we capture an initial RGB-D frame and apply GroundingDINO~\cite{liu2023grounding} to segment an RGB mask of the object. We use the real world object's center point (determined by averaging the depth values within the RGB mask) to initialize the object's $x-$ and $y-$coordinates in simulation.

\textbf{Data Collection Pipeline.} 
Using the teleoperation pipeline described in Sec.~\ref{sec:setup}, we collect four source human demonstrations for the Can Sorting task. These demonstrations are replayed in simulation, and are used as source demonstrations for \sysName in the digital twin.
Next, new real-world demonstrations are collected by synchronizing the initial object state from real to sim, and then attempting to generate a new demonstration in sim with \sysName. If the demonstration is successful in simulation, the sequence of robot control actions is sent to the real-world for execution.
In this way, the digital twin functions to ensure safety during real-world data generation, while \sysName mitigates human effort for data collection, which is autonomous apart from the environment resets.
We generate 40 successful demonstrations with the approach described above.%

\textbf{Results.} 
We compare visuomotor policies trained using Diffusion Policy \cite{chi2023diffusion} on the 40 \sysName demos with one trained on the 4 source demos.
We evaluated both models by running 10 trials each for the red and blue cups. 
The policy trained on \sysName data achieves $90\%$ success, while the model trained on the source data achieves $0\%$; \sysName thus offers an efficient pipeline for training real-world robots through the use of a digital twin.

\section{Conclusion}
\label{sec:conclusion}

We introduce \sysName, a large-scale automated data generation system that synthesizes trajectories from a small number of human demonstrations for bimanual and dexterous robots, and a collection of nine simulation environments across three embodiments requiring different coordination behaviors. Our findings from applying \sysName to these tasks show that there is great value in further investigating policy learning in this setting. We also deploy DexMimicGen on a real humanoid robot through a real2sim2real pipeline. We hope the release of our \sysName datasets and environments will facilitate future research.

\section*{ACKNOWLEDGMENT}
We appreciate Fourier Intelligence for hardware support. We also thank Yifeng Zhu, Abhiram Maddukuri, Soroush Nasiriany, and Yu Fang for their help with robosuite, and Akul Santhosh and Abhishek Joshi for their help with rendering, and Toru Lin and Tairan He for valuable discussions.

\clearpage

\bibliographystyle{IEEEtran}
\bibliography{IEEEabrv,dexmg}

\newpage
\section{Appendix Overview}
\label{app:overview}

The Appendix contains the following content.

\begin{itemize}
    \item \textbf{Implementation Details} (Appendix~\ref{app:implementation}): more details of DexMimicGen implementation.
    \item \textbf{Result Analysis} (Appendix~\ref{app:analysis}): analysis of DexMimicGen results.
    
    \item \textbf{Author Contributions} (Appendix~\ref{app:author}): list of each author's contributions to the paper.
\end{itemize}

\section{Implementation Details}
\label{app:implementation}

\noindent\textbf{Which parts of the DexMimicGen process rely on human input versus automation?}  
\begin{itemize}
    \item The source demonstration collection requires human teleoperation.
    \item Similar to MimicGen, we have two options for segmenting the source demonstrations. The first option relies on manually defined heuristics, where we implement subtask terminal signals --- e.g., detecting when the hand makes contact with the target object in simulation --- and automatically segment the source demonstrations by checking the corresponding simulation states. The second option involves manually segmenting each demonstration, which requires more human effort but offers greater flexibility, especially when subtask terminal signals are difficult to define.
    \item By default all subtasks are parallel subtasks. We need to manually specify which pairs of subtasks are coordination subtasks or sequential subtasks if required.
\end{itemize}

Once the source demonstrations are collected and segmented, and the subtask structure is specified, the data generation process is fully automated.

\vspace{1mm}
\noindent\textbf{How does DexMimicGen determine the success condition of a task?}  
We implement a success check function for each task. Typically, success is determined based on the final simulation state, such as whether the object of interest is placed in the target container. The success check is used for filtering out failed demonstrations during the data generation phase.

\vspace{1mm}
\noindent\textbf{How does DexMimicGen handle collisions between the robots and objects?}  
DexMimicGen does not explicitly handle collisions. Some failure cases during the data generation phase result from collisions between the generated trajectory and objects in the workspace. To mitigate this issue, we plan to extend DexMimicGen with motion planning modules from SkillMimicGen~\cite{garrett2024skillmimicgen} for future work.

\section{Result Analysis}
\label{app:analysis}

\noindent\textbf{What factors contributed to the low success rate of certain tasks?}  
For instance, the threading task has a success rate below 70\%. We hypothesize that in this task, both the threading object and the hole are occluded from the third-person camera, making it challenging for the vision-based policy to complete the task successfully. To address this issue, we could incorporate visual reinforcement learning to enable active perception and improve dexterous control. We believe it will facilitate the policies to accomplish the tasks under high occlusions.

\begin{figure}
    \centering
    \includegraphics[width=\linewidth]{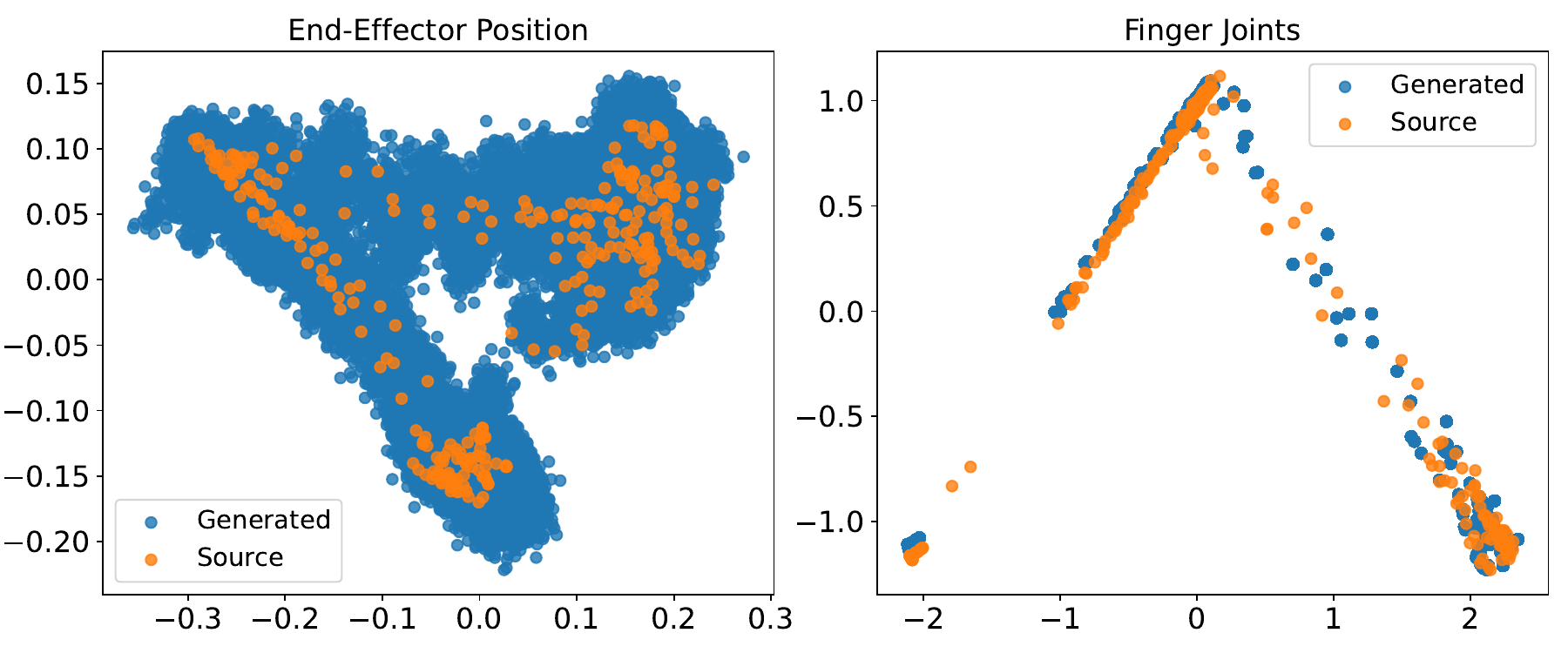}
    \caption{\textbf{Visualization of generated and source action distributions.} We run PCA to project actions into 2D and visualize them.}
    \label{fig:pca-vis}
\end{figure}

\noindent\textbf{How does the DexMimicGen process augment the data distribution?}  
To further analyze the data generation process of DexMimicGen, we visualize the PCA projections of end-effector poses and finger joint actions for both generated and source demonstrations in the TwoArmCoffee task (Fig.~\ref{fig:pca-vis}). The results show that DexMimicGen significantly expands the distribution coverage of end-effector actions. In contrast, for finger joint actions, DexMimicGen primarily performs local interpolation rather than broad expansion.

\section{Author Contributions}
\label{app:author}

\textbf{Zhenyu Jiang.} Co-led project ideation and development. Implemented the data generation code and simulation environments. Oversaw the development of the teleoperation and control infrastructure. Ran most of the experiments in the paper, and wrote the paper.

\textbf{Yuqi Xie.} Core developer of the project. Developed the simulation environments, teleoperation infrastructure for simulation, and rendering pipeline. Ran part of the experiments for humanoids, and the real robot experiments.

\textbf{Kevin Lin.} Core developer of the project. Developed the control infrastructure for the simulation experiments, including whole-body IK controllers. Ran part of the experiments for humanoids.

\textbf{Zhenjia Xu.} Implemented the real robot teleoperation and policy deployment infrastructure and helped oversee the real robot experiments.

\textbf{Weikang Wan.} Implemented the initial prototype of the data generation code and ran the BiGym~\cite{chernyadev2024bigym} experiments.

\textbf{Ajay Mandlekar.} Co-led project ideation and development. Implemented simulation environments. Oversaw the development of the main algorithm for data generation, the simulation environments, and the experiments presented in the paper. Advised on the project and wrote the paper.

\textbf{Linxi Fan.} Co-led project ideation and development. Led resource acquisition for the project, including robot hardware and cluster compute. Provided feedback on paper writing.

\textbf{Yuke Zhu.} Co-led project ideation and development. Provided feedback on experiments and presentation, and wrote the paper.

\end{document}